\pdfoutput=1
\documentclass[11pt]{article}

\usepackage[final]{acl}

\usepackage{times}
\usepackage{latexsym}
\usepackage{booktabs}
\usepackage{amsmath}
\usepackage{enumitem}
\usepackage{colortbl}
\usepackage{hyperref} % 用于超链接
\usepackage{cleveref} % 让 cleveref 自动处理引用前缀
\usepackage{tcolorbox}
\usepackage{algpseudocode} 
\usepackage{algorithm}
\usepackage{xspace}
\usepackage[T1]{fontenc}
\usepackage[utf8]{inputenc}

\usepackage{microtype}

\usepackage{inconsolata}
\usepackage{pifont}
\definecolor{Maroon}{rgb}{0.5, 0, 0} 
\definecolor{OliveGreen}{rgb}{0.33, 0.42, 0.18}
\definecolor{Violet}{cmyk}{0.25, 0.5, 0, 0}
\newcommand{\cmark}{\color{OliveGreen}{\ding{51}}}%
\newcommand{\xmark}{\color{Maroon}{\ding{55}}}%
\usepackage{graphicx}
\usepackage{multirow}
\newcommand\ours{\textsc{AutoReproduce}\xspace}
\newcommand\ourbench{\textsc{ReproduceBench}\xspace}

\title{\textsc{AutoReproduce}: Automatic AI Experiment Reproduction with Paper Lineage}

\author{%
Xuanle Zhao$^\text{1}$,\enspace Zilin Sang$^\text{2}$,\enspace Yuxuan Li$^\text{1,}$\footnotemark[2],\enspace Qi Shi$^\text{1,}$\footnotemark[2],\enspace Weilun Zhao$^\text{3}$,\enspace Shuo Wang$^\text{1}$,\\ \textbf{Duzhen Zhang$^\text{4}$,\enspace Xu Han$^\text{1}$,\enspace Zhiyuan Liu$^\text{1}$,\enspace Maosong Sun$^\text{1}$} \\
\textsuperscript{1}Tsinghua University
\textsuperscript{2}Xidian University \\
\textsuperscript{3}OpenBMB \textsuperscript{4}University of the Chinese Academy of Sciences\\[2pt]
\texttt{\{2429527z, yxuanl1995, qshi9510\}@gmail.com}
}

\begin{document}
\maketitle
\renewcommand{\thefootnote}{\fnsymbol{footnote}}
\footnotetext[2]{Corresponding authors.}
% \footnotetext[2]{Corresponding author.}
\renewcommand{\thefootnote}{\arabic{footnote}}

\begin{abstract}
Efficient reproduction of research papers is pivotal to accelerating scientific progress. However, the increasing complexity of proposed methods often renders reproduction a labor-intensive endeavor, necessitating profound domain expertise.
To address this, we introduce the paper lineage, which systematically mines implicit knowledge from the cited literature. This algorithm serves as the backbone of our proposed \ours, a multi-agent framework designed to autonomously reproduce experimental code in a complete, end-to-end manner. To ensure code executability, \ours incorporates a sampling-based unit testing strategy for rapid validation. To assess reproduction capabilities, we introduce \ourbench, a benchmark featuring verified implementations, alongside comprehensive metrics for evaluating both reproduction and execution fidelity. Extensive evaluations on PaperBench and \ourbench demonstrate that \ours consistently surpasses existing baselines across all metrics. Notably, it yields substantial improvements in reproduction fidelity and final execution performance. The code is available at \url{https://github.com/AI9Stars/AutoReproduce}.
% The code and benchmark will be released soon.
% In particular, compared to the official implementations, \ours achieves an average performance gap of $17.1\%$ on $94.9\%$ of the executable experiment runs.

\end{abstract}

% \renewcommand{\thefootnote}{\fnsymbol{footnote}}
% \footnotetext[2]{Corresponding authors.}
% % \footnotetext[2]{Corresponding author.}
% \renewcommand{\thefootnote}{\arabic{footnote}}

\section{Introduction}
The rapid advancement of artificial intelligence (AI) has heightened the demand for efficient workflow iterations, making the ability to reproduce experimental results increasingly critical \cite{xi2025rise, li2025system}. 
While this capability is pivotal for advancing various fields, the complexity of method designs and training pipelines, often requiring specialized expertise for comprehension \cite{si2024can, li2024chain}, substantially hinders automated experimental replication.
For instance, developing a task-specific model often requires collaboration among experts in data processing, model design, and training pipeline \cite{qian2023chatdev}.

Notably, the rapid evolution of AI has resulted in a substantial corpus of research papers, presenting a valuable testbed for exploring the automation of experiment replication \cite{erdil2023explosive}.
Prior work has typically focused utilizing large language models (LLMs) for automating in-depth analyses of existing papers (e.g., report or idea generation \cite{baek2024researchagent}) or assisting with discrete reproduction tasks (e.g., environment setup \cite{bogin2024super}, code repository refactor \cite{gandhi2025researchcodeagent, siegel2024core}). However, comprehensive frameworks for automatic end-to-end reproduction remain unexplored.
While the concurrent work \cite{seo2025paper2code} also attempts to address the task of automatic reproduction. They only focus on reproducing the contents introduced in the paper, without considering the executability of the generated code, which is crucial for faithful experimental reproduction \cite{wang2024executable}.

Reproducing experiments efficiently remains a significant challenge due to insufficient experimental details in research papers. We observe that distinct research domains often rely on tacit knowledge, that is, common implementation practices like specific module architectures \cite{chen2024transunet} and data processing pipelines \cite{nie2022time} that evolve into de facto standards for subsequent studies. An effective reproduction strategy must incorporate the domain-specific knowledge and practices surrounding the source paper.

Informed by the preceding analysis, we propose the paper lineage algorithm, which identifies potentially unstated details by tracing cited literature and associated code repositories of the source paper. Building upon this algorithm, we propose \ours, a multi-agent framework designed for the end-to-end reproduction of experiments in papers. 
\ours contains three key stages, \textit{literature review}, \textit{paper lineage} and \textit{code development}, designed to be executed sequentially to generate valid reproductions. 
During code development, we propose a sampling-based unit testing strategy for rapid validation, thereby guaranteeing the code executability.

Furthermore, to rigorously evaluate the efficacy of \ours, we curate \ourbench, a benchmark comprising 13 research papers that span distinct AI sub-domains. For each entry, we manually construct and execute verified implementations to establish ground-truth performance.
In our setup, LLM agents are tasked with reproducing the specific experiment implementations detailed in the source papers. To assess the generated code, we employ five distinct metrics spanning dimensions from structural reproduction fidelity to final execution accuracy. Experimental results demonstrate that \ours achieves superior performance on \ourbench across all five metrics, validating the effectiveness of our proposed approach. Our main contributions are summarized as follows:

\begin{itemize}[leftmargin=*]
    \item We propose \textit{paper lineage}, an algorithm that enables the agent to learn implicit domain knowledge and implementation practices by analyzing the cited literature.
    \item We propose \ours, a multi-agent framework designed for end-to-end experiment reproduction, which achieves superior performance compared with existing methods.
    \item We construct \ourbench, a novel benchmark for evaluating the experiment reproduction capabilities,  which includes manually curated reference code implementations and a suite of evaluation metrics.
\end{itemize}

\section{Related Work}
\subsection{LLMs for Experiment Automation}

Large Language Models (LLMs) are increasingly utilized to automate diverse stages within machine learning workflows, including data engineering, model selection, hyperparameter optimization, and workflow evaluation \cite{gu2024large}.
For instance, in data engineering, LLMs are utilized to assist with many tasks such as dataset recommendation \cite{yang2025autommlab}, adaptive data imputation \cite{zhang2023automl}, context-aware data transformation \cite{liu2023jarvix}, and feature selection \cite{jeong2024llm}. Also, many works explore utilizing LLM-driven approaches for model selection, for example, $\text{AutoM}^3\text{L}$ \cite{luo2024autom3l} proposes a retrieval-based process to select the required model, while ModelGPT \cite{tang2024modelgpt} employs generation-based methods for the same purpose. Furthermore, LLMs contribute to workflow evaluation by enabling performance prediction \cite{zhang2023automl} and supporting zero-cost proxy method development \cite{hollmann2022tabpfn}.

% While current works automate discrete experimental stages, their lack of comprehensive end-to-end automation limits utility and scalability for large-scale workflows. In contrast, \ours aims to provide a full end-to-end automation of the experimental workflow.

\subsection{LLMs for Research Code}
Prior works have explored using LLMs for generating novel ideas \cite{li2024chain, weng2024cycleresearcher}. For example, Scimon \cite{wang2024scimon} focuses on scientific hypothesis discovery by elucidating relationships between variables, while ResearchAgent \cite{baek2024researchagent} employs an agent-based system to generate open-ended ideas accompanied by conceptual methods and experimental designs. However, these approaches typically do not generate implementation code for these novel concepts, leading to unverifiable results.
More recently, many works \cite{schmidgall2025agent, schmidgall2025agentrxiv} utilize multi-agent frameworks to generate code implementations of proposed ideas, further promoting this field. Despite this advancement, the ideas generated by these approaches often remain paper-level concepts, lacking the detailed substance described in formally proposed papers.

Achieving comparable performance in experimental reproduction hinges on the high-fidelity implementation of the methods detailed in the source paper. Recent concurrent studies \cite{starace2025paperbench, seo2025paper2code, lin2025autop2cllmbasedagentframework} further underscore the significance of this research domain.

\section{\ours}
\begin{figure*}
  \centering
  \includegraphics[width=0.97\linewidth]{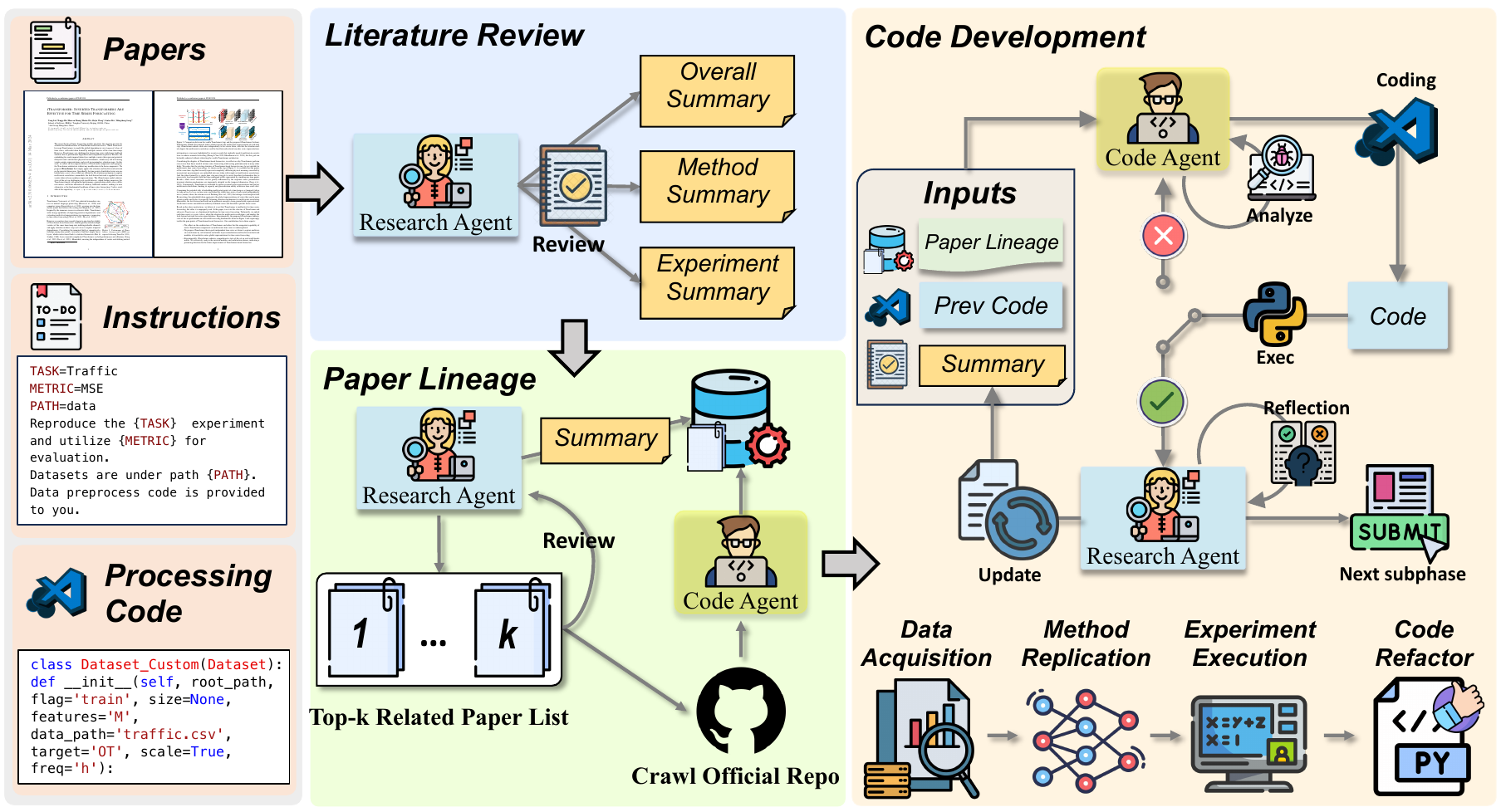}
  \vspace{-5pt}
  \caption{The paper content, instructions and data processing code (if necessary) are provided for each reproduction task. The workflow of \ours, which is decomposed into three subphases. (i) Literature Review, the research agent summarize the overall, method and experiment contents. (ii) Paper Lineage, the research agent lists and reviews related papers, and the code agent filters files in the corresponding repositories. (iii) Code Development, the code agent and research agent collaborate to construct executable reproduction code.}
  \vspace{-10pt}
  \label{fig:main}
\end{figure*}

\subsection{Problem Formulation}
Rapid AI progress yields numerous novel methods, yet manual reproduction imposes prohibitive costs in time and expertise. We thus define automated experiment reproduction as the task of utilizing LLM agents to generate executable code for experiment replication, thereby automating scientific verification. Given a paper $\mathcal{P}$ and instructions about experiment $\mathcal{I}$, the agent $\mathcal{A}$ needs to reproduce the code implementation of the method and experiment proposed in the paper $\mathcal{C}=\mathcal{A}(\mathcal{P}, \mathcal{I})$, where $\mathcal{C}$ is the output code.

\subsection{Workflow}
We introduce \ours, a novel multi-agent framework for reproducing experiments from research papers. As illustrated in Figure~\ref{fig:main}, its pipeline is structured into three key phases: (i) Literature Review, (ii) Paper Lineage, and (iii) Code Development. This process is collaboratively executed by two specialized agents: a research agent for text-centric tasks such as paper summarization and related work analysis, and a code agent responsible for all code-oriented tasks, including implementation and debugging.

\subsubsection{Literature Review}
To mitigate the information redundancy inherent in research papers, the Research Agent employs a hierarchical three-stage summarization protocol to comprehensively extract core methodologies and experimental nuances. The process begins with a holistic paper-level overview, followed by targeted summaries that distill specific method details, such as mathematical formulations and implementation specifics, as well as the critical experimental settings required for reproduction.
Acknowledging that effective summarization hinges on the quality of text extraction, where direct parsing methods \cite{schmidgall2025agent} often struggle with complex artifacts like mathematical formulas and tables, we employ MinerU \cite{wang2024mineru}. This tool converts PDFs into Markdown format, significantly enhancing the fidelity of data preservation. Furthermore, to augment semantic understanding, \ours offers an optional capability to enrich summaries by integrating insights from visual structure diagrams.

\subsubsection{Paper Lineage}
Scientific research is an inherently cumulative process, where novel methods evolve upon the foundation of existing studies. This iterative progression fosters domain-specific conventions and implicit consensus, forming a historical context we term the \textit{Paper Lineage}. To systematically exploit this, we propose the Paper Lineage algorithm to analyze the research landscape and uncover these prevailing practices.
Specifically, the research agent identifies the top-$k$ relevant papers (default $k$=3) from the source paper's references. This selection is driven by an analysis of citation relationships within the source paper's full context, where relevance is defined by the alignment of research fields and proposed methods. Notably, comparison baselines detailed in the primary experimental section are prioritized as the most critical references for analysis. Upon acquiring the relevant paper list, the research agent retrieves the manuscripts via the \texttt{ArXiv API}, summarizes their content, and identifies linked code repositories. It then employs the \texttt{GitHub API} to clone the corresponding repository. To isolate relevant files from extraneous repository content, the code agent leverages the paper summary and task instructions to selectively extract essential source files. These code segments are paired with their summaries to construct \textit{<summary, code>} tuples, which serve as domain-aligned reference exemplars for subsequent generation. For papers lacking public repositories, their summaries alone are utilized as high-quality conceptual knowledge sources. The detailed procedure and prompts are outlined in Algorithm \ref{alg:paper_lineage} and Figure \ref{fig:paper_lineage_prompt_research_agent}, \ref{fig:paper_lineage_prompt_code_agent}.
% The Algorithm~\ref{alg:paper_lineage} summarizes the Paper Lineage framework.

\subsubsection{Code Development}
The code development phase constitutes the final stage of the \ours workflow, where the research and code agents collaborate to produce a reproducible and executable implementation of the target experiment. To guarantee a high-fidelity reproduced implementation, this phase is structured into three main stages, culminating in a final refactoring of the generated code. To facilitate code execution and debugging, we provide a Docker container that encapsulates the Python runtime and common libraries such as PyTorch and Numpy. When encountering missing dependencies, the code agent can utilize \texttt{\textasciigrave\textasciigrave\textasciigrave Bash\textbackslash n<command>\textbackslash n\textasciigrave\textasciigrave\textasciigrave} to install the necessary packages.

\textbf{Data Acquisition.} 
The preliminary phase focuses on dataset curation and attribute analysis. Although off-the-shelf datasets from libraries like \texttt{torchvision} are readily available, many tasks demand the curation of custom datasets derived directly from raw data.
Reproducing prior work is often impeded by the insufficient documentation of data preprocessing method and the absence of standardized data formats.
To mitigate this issue, we categorize source papers based on dataset provenance: standard benchmarks or custom datasets. For papers relying on standard benchmarks, \ours generates loading code using established libraries. For those involving custom datasets, we provide corresponding preprocessing pipelines to enable seamless agent utilization.
Furthermore, to prevent runtime failures arising from mismatches in critical data properties, such as tensor shape and data type, \ours employs a proactive inference mechanism. The code agent extracts these attributes by generating and executing analysis code on sampled mini-batches, thereby gathering essential context to guide the subsequent code generation process.

In response to execution errors, the code agent first diagnoses the error traceback and refines the script using the \texttt{EDIT} command, structured as \texttt{\textasciigrave\textasciigrave\textasciigrave EDIT\textbackslash n N M\textbackslash n<new code>\textbackslash n\textasciigrave\textasciigrave\textasciigrave}. We decouple error analysis and code editing into two distinct steps, as we observe that conducting a preliminary analysis to guide the debugging process significantly improves the success rate.
The \texttt{EDIT} command facilitates targeted updates by replacing lines \texttt{N} through \texttt{M} with the generated code segment, rather than regenerating the entire file. This granular approach significantly reduces token generation overhead and is employed consistently throughout all subsequent phases of our framework. The prompt of the command is illustrated in Figure \ref{fig:edit_prompt}.

\begin{table*}[t]
\centering
\begin{small}
\setlength{\tabcolsep}{15pt}
\begin{tabular}{lllc}
\toprule
\textbf{Method} & \textbf{Domain} & \textbf{Dataset} & \textbf{Metrics} \\
\midrule
\emph{IEBins}~\cite{shao2023iebins} & Monocular Depth Estimation & NYU-Depth-v2
    & $\delta < 1.25$\\
\emph{iTransformer}~\cite{liu2023itransformer} & Time Series Forecasting & Traffic & MSE\\    
\emph{DKD}~\cite{zhao2022decoupled} & Knowledge Distillation & CIFAR-100 & Accuracy\\
\emph{SimVP}~\cite{gao2022simvp} & Video Prediction & Moving MNIST & MSE\\
\emph{HumanMAC}~\cite{chen2023humanmac} & Human Motion Prediction & HumanEva-I & ADE\\
\emph{SFNet}~\cite{cui2023selective} & Image Dehazing & SOTS-Indoor & PSNR\\
\emph{LSM}~\cite{wu2023solving} & Solving PDEs & Darcy & MSE\\
\emph{Swin-Unet}~\cite{cao2022swin} & Medical Image Segmentation & Synapse & DSC\\
\emph{TDGNN-w}~\cite{wang2021tree} & Node Classification & Citeseer & Accuracy\\
% \emph{NBFNet}~\cite{zhu2021neural} & Link Prediction & WN18RR & H@10\\
\emph{TimeVAE}~\cite{desai2021timevae} & Time Series Generation & Sine 20\% & Predictor\\
\emph{WCDM}~\cite{jiang2023low} & Low-light Image Enhance & LOLv1 & SSIM\\
\emph{BSPM}~\cite{choi2023blurring} & Collaborative Filtering & Gowalla & Recall\\
\emph{DAT-S}~\cite{chen2023dual} & Image Super-Resolution  & DF2K/Set5  & PSNR\\
\bottomrule
\end{tabular}
\end{small}
\vspace{-5pt}
\caption{List of papers in \ourbench. To illustrate the specific experiments, we list each experimental detail, including the method names, task domains, datasets, and evaluation metrics.}
\label{tab:paper_list}
\vspace{-10pt}
\end{table*}

\textbf{Method Replication.}
This stage is dedicated to implementing the method. The code agent synthesizes code snippets and resolves errors by leveraging established context, including the paper summary, inferred data attributes, and domain knowledge retrieved from the Paper Lineage. Simultaneously, the research agent validates the generated code against the method summary, providing corrective feedback while dynamically updating the summary to steer the implementation.
The code agent begins by generating an initial implementation, which is then iteratively refined through a collaborative mechanism. While the code agent debugs potential errors in model computations by inspecting data flow properties, the research agent validates the code against the paper summary, updating the summary to guide the code agent in resolving any identified discrepancies
The research agent chooses to submit the code once it fully aligns with the paper.
% After this subphase, the method proposed in the paper is reproduced and ready for subsequent experiments.

\textbf{Experiments Execution}
This phase is dedicated to the implementation of the full experimental pipeline. Leveraging the established dual-agent loop, the system verifies that the code correctly reflects the experimental settings from the source paper. We continue to employ mini-batch sampling to accelerate debugging.
To validate epoch-related configurations, the code agent generates the complete experiment script equipped with early-exit mechanisms, such as \texttt{break}. This enables a rapid dry run to validate the pipeline prior to full training.

This phase concludes with a comprehensive refactoring process to remove unnecessary debug settings and clean up the generated code.

\section{\ourbench}

\subsection{Contributions}
% Paper replication has recently garnered significant research attention, leading to the proposal of several benchmarks and methods. 
Our approach prioritizes experiment reproducibility by focusing on code execution and final performance. This distinguishes our work from methods like Paper2Code \cite{seo2025paper2code}, which generate code in a single pass without considering execution. To this end, we construct \ourbench, a paper reproduction benchmark that establishes a comprehensive evaluation framework to assess agent fidelity and performance.

\subsection{Paper Selection}
We establish a rigorous curation pipeline utilizing PapersWithCode, covering diverse domains such as CV and NLP along with their respective sub-domains. Within each category, we screen 5–10 candidate papers to verify code completeness and reproducibility, culminating in the selection of at most one high-quality representative per sub-domain. Consequently, \ourbench comprises 13 human-curated papers spanning a broad spectrum of complexities, ranging from foundational applications like knowledge distillation \cite{zhao2022decoupled} to specialized challenges such as solving Partial Differential Equations (PDEs) \cite{wu2023solving}. Moreover, these papers encompass varied experimental conditions, including the two primary paradigms of training from scratch and fine-tuning pre-trained models. The construction and validation of \ourbench follow a three-stage protocol. First, we identify representative method variants, experimental settings, and evaluation metrics specific to each paper. Next, we manually curate the official code repositories. This critical step involves excising boilerplate and refactoring the repository to isolate the core implementation from auxiliary logic. Finally, we re-execute all experiments to validate reproducibility and establish a verified baseline, ensuring fair and unambiguous evaluation. Detailed statistics are listed in Table~\ref{tab:paper_list}. We adopt these reproduced results as the ground truth to guarantee consistent benchmarking and mitigate potential misinterpretation.

\subsection{Evaluation}
Given the availability of reference code and performance baselines in our curated \ourbench, we conduct evaluations from two key perspectives to explore: \textit{(i) Does the generated code accurately reflect the core contributions and experimental setup as proposed in the source paper?} \textit{(ii) Can the generated code fully reproduce the experimental performance metrics rerun by ourselves?}
To answer these questions, we introduce two primary evaluation metrics, evaluating from alignment and execution aspects, respectively.
\begin{table*}[t]
\centering
\setlength{\tabcolsep}{2pt}
\begin{tabular}{l|c|ccc|cc}
\toprule
\multirow{2}{*}{Baselines} & \multirow{2}{*}{LLM} & \multicolumn{3}{c|}{\textbf{Align-Score}}  &  \multicolumn{2}{c}{\textbf{Exec-Score (\%)}}  \\
\cmidrule{3-7} 
 &  & Paper-Level & Code-Level & Mixed-Level & Exec Rate & Perf Gap ($\downarrow$) \\
\toprule
ChatDev & GPT-4o & $57.33$ & $32.80$ & $43.33$ & $2.56$ & $99.62$ \\
Agent Laboratory & GPT-4o & $63.47$ & $35.32$ & $48.64$ & $23.08$ & $82.31$ \\
% MLAgent & GPT-4o &  \\
PaperCoder & o3-mini  & $90.41$ & $47.54$ & $60.26$ & $17.94$ & $89.23$\\ 
\midrule
\multirow{4}{1.5cm}{\textsc{Auto-}\\\textsc{Reproduce}}
 & GPT-4o & $82.13$ & $41.52$ & $56.24$ & $76.92$ & $41.77$ \\
 & \scalebox{0.95}{Claude-3.5-Sonnet}  & $90.27$ & $54.11$ & $69.97$ & $84.62$ & $31.62$\\
 & o3-mini & $90.86$ & $58.48$ & $75.21$ & $92.31$ & $24.31$\\
& \scalebox{0.95}{Gemini-2.5-Pro} & $\mathbf{91.57}$ & $\mathbf{60.26}$ & $\mathbf{77.56}$ & $\mathbf{94.87}$ & $\mathbf{19.72}$\\
\bottomrule 
\end{tabular}
\vspace{-5pt}
\caption{The evaluation of various agents on \ourbench, utilizing both o1-as-judge and execution. The presented results for each metric represent the mean value derived from three independent runs conducted across all papers in the benchmark. The best performance is indicated in \textbf{Bold}. }
\label{tab:overall_results}
\label{tab:main_results}
\vspace{-10pt}
\end{table*}

\textbf{Align-Score:} 
To assess alignment fidelity, we conduct a comparative analysis bridging the generated code, the core content of the paper, and the manually curated reference implementation. We examine the results across three distinct dimensions. 
(1) Paper-Level: Given the source paper, we utilize an LLM (default \texttt{o1}) to extract five critical components essential for experimental reproduction. Subsequently, we prompt the LLM to assess the degree to which the generated code satisfies these key objectives.
(2) Code-Level: Leveraging manually annotated reference implementations, we first ensure that extraneous logic is stripped away to establish a clean ground truth. We then prompt the LLM to evaluate the generated code against this reference across four specific dimensions, encompassing overall structure, model details, training details, and experimental integrity.
3) Mixed-Level: Our analysis reveals that paper-level evaluation often overlooks granular implementation details, potentially inflating scores, whereas code-level evaluation can be overly sensitive to syntactic variations, leading to score deflation. To address this trade-off, we propose the mixed-level evaluation strategy. This method supplies the LLM with both the key objectives extracted from the paper and the reference code context. Our evaluation results demonstrate that enabling the model to ground abstract requirements in concrete implementation patterns leads to significantly more nuanced scoring of the generated code. Consequently, this approach captures both critical features and detailed logic, demonstrating improved consistency with human judgment. Unlike Paper2Code \cite{seo2025paper2code}, which relies on a single holistic score (1-5 scores), our framework offers a fine-grained assessment by validating individual implementation points.

\textbf{Exec-Score:} Given that experimental reproduction is fundamentally a code generation task, the execution outcomes of the generated code are of paramount importance. We assess this dimension by measuring the Execution Rate (Exec Rate) and the final experimental performance gap (Perf Gap). To address the heterogeneity of performance metrics across diverse research papers, we propose the relative performance gap as a unified evaluation standard. This metric quantifies the relative deviation between the final results yielded by the generated code and the reference performance established by our curated implementations.
% Given that evaluation metrics vary across methods, we assess agent performance by comparing their average relative metric gap,
\begin{equation}
\small
\text{Performance Gap}=\frac{1}{n}\sum_{i=1}^n\frac{|\text{P}^\text{ref}_i-\text{P}^\text{agent}_i|}{\max(\text{P}^\text{ref}_i, \text{P}^\text{agent}_i)}
\end{equation}
where $\text{P}_{ref}$ and $\text{P}_{agent}$ are the performance obtained under the reference code and agent-generated code, respectively. 
Since not all generated code can be executed, the $\text{P}_{\text{agent}, i}$ is set to 0 for the non-executable instances, resulting in a maximum performance gap of 1.0.
Furthermore, to prevent the performance gap from exceeding 1.0, especially in cases with small reference performance values (e.g. $\text{P}^\text{ref}_i=0.1$ and $\text{P}^\text{agent}_i=0.3$ when utilizing MSE as the evaluation metric), we normalize the gap by the larger performance value of $\text{P}^\text{ref}$ and $\text{P}^\text{agent}$.
Reference performance values are derived by executing our verified ground-truth implementations.

\section{Experiments}
\subsection{Baselines and Benchmarks}
% Prior research predominantly centres on software development \cite{qian2023chatdev} rather than the requirements of experiment reproduction. While both approaches leverage code to achieve specified goals, their core implementation principles differ substantially. Reproducing experiments necessitates a rigorous correspondence between the implemented code and the method detailed in the source paper, whereas software development often prioritizes overall system functionality and completeness.
% Meanwhile, prior works have explored agent-assisted paper generation \cite{schmidgall2025agent}, which typically demonstrate their proposed innovations by generating associated code and authoring academic papers. However, a key limitation is that these innovations, often abstract, lack implementation details, thereby hindering generate concrete implementations \cite{baek2024researchagent}.
To better compare our proposed \ours, we compare previous work on software development and paper generation, including ChatDev \cite{qian2023chatdev}, Agent Laboratory \cite{schmidgall2025agent} and PaperCoder \cite{seo2025paper2code}. To our knowledge, PaperCoder is concurrent work compared to \ours. For all the metrics in the align-score, we employ \texttt{o1} as the LLM judge for evaluations. 
Beyond \ourbench, we also evaluate \ours on PaperBench Code-Dev \cite{starace2025paperbench}, which imposes requirements distinct from our primary benchmark. To align with its focus on static code generation for all experiments without execution, we streamline the \ours by omitting the iterative debugging phase and configuring the system to generate the complete experimental code including baseline methods.

\subsection{Main Results}
We evaluate the implementation code generated by various baselines using different LLM backbones, with three runs conducted for each paper. Detailed results in Table \ref{tab:overall_results} demonstrate that our proposed \ours achieves superior performance across all the metrics, notably in execution rate and performance gap. By employing a sampling batch approach for efficient debugging, \ours significantly enhances the execution rate of the generated code and reduces its performance gap with the reference code. 
In addition, as shown in Table \ref{tab:results_paperbench}, \ours also achieves superior performance on the PaperBench Code-Dev benchmark, particularly when utilizing the paper lineage algorithm. We also observe that enabling the agents to process information from visual diagrams further improves performance, as some experimental settings are denoted within the diagrams.

Our analysis indicates that LLM judges may overrate consistency when comparing paper contents against generated code. This overestimation stems from the generality of textual descriptions, which can reward broad functional similarity rather than precise replication. Conversely, directly comparing generated code with reference implementations is also problematic, as reference code includes settings unmentioned in the paper, thereby skewing reproducibility assessments. In contrast, our proposed mixed-level scoring aligns more closely with human evaluators and offers a more reliable metric for evaluating reproduction capabilities.

% Moreover, our analysis indicates that when using LLMs as judges, solely comparing key points of the paper content and the generated code tends to overrate consistency. This overrating of consistency by LLM judges arises from the inherent generality of the textual descriptions, causing high scores for broad functional code similarity.
% However, directly comparing generated and reference code is problematic because the reference often includes settings not mentioned in the paper, which skews the reproducibility assessment. In contrast, the mixed-level scoring aligns more closely with human evaluation and provides a more reliable metric for assessing the reproducing capacity of the LLM agent.
\begin{table}[t]
% \small
\centering
\setlength{\tabcolsep}{3pt}
\begin{tabular}{lcc}
\toprule
Baselines & Backbone & Rep. Score (\%)  \\
\toprule
BasicAgent & o3-mini & $6.4$ \\
IterativeAgent & o3-mini & $17.3$ \\
IterativeAgent & o1-high & $43.4$ \\
PaperCoder & o3-mini & $45.1$ \\
\midrule
\ours \\
(w/o Paper Lineage) & o3-mini & $44.1$ \\
Default Setting & o3-mini & $48.5$ \\
(w/ Visual Diagram) & o3-mini & $49.6$ \\
\bottomrule 
\end{tabular}
\vspace{-5pt}
\caption{valuation results on PaperBench Code-dev are presented. Rep. Score denotes the Replication Score. The final three rows represent various configurations of \ours.}
\label{tab:results_paperbench}
\vspace{-10pt}
\end{table}

\subsection{Ablation Study}
To explore and evaluate our proposed methods, we conduct ablation experiments to investigate: (i) the utility of the visual diagrams and the impact of content extraction MinerU during literature review, (ii) the effectiveness of our proposed Paper Lineage method and (iii) the performance difference between including debugging and refinement versus one-time code generation in the Code Development phase. Due to time constraints, we conduct only an experiment for each paper and evaluate on the Mixed-Level and Performance Gap metrics. 

\begin{table}[t]
% \small
\centering
\setlength{\tabcolsep}{2pt}
\begin{tabular}{lcc}
\toprule
Ablations & Mixed-Level & Perf Gap \% ($\downarrow$) \\
\midrule
\textit{w/} Visual Diagram & $70.14$ & $35.83$ \\
\textit{w/o} MinerU  & $58.42$ & $47.81$\\
\midrule
\textit{w/o} Paper Lineage  & $63.15$ & $39.59$ \\
\midrule
\textit{w/o} Refine & $65.78$ & $36.37$\\
\textit{w/o} Debug+Refine & $68.32$ & $88.78$\\
\midrule
\ours & $69.97$ & $31.62$ \\
\bottomrule
\end{tabular}
\vspace{-5pt}
\caption{The ablation study for evaluating \ours with various subphases, the results are measured using mixed-level and performance gap metrics. We employ \texttt{Claude-3.5-Sonnet} as the LLM backbone.}
\label{tab:ablation}
\vspace{-10pt}
\end{table}

\begin{table*}[t]
% \small
\centering
\setlength{\tabcolsep}{6pt}
\begin{tabular}{l|c|cccc}
\toprule
Baselines & LLM & Method & Parameter & Experiment & Overall \\
\midrule
ChatDev & GPT-4o & $4.08\pm1.00$ & $2.85\pm0.37$ & $1.92\pm0.15$ & $8.86\pm1.12$ \\
PaperCoder & o3-mini & $6.84\pm0.52$ & $3.46\pm0.31$ & $2.92\pm0.23$ & $13.24\pm0.68$ \\
\scalebox{0.95}{\ours} & \scalebox{0.95}{Claude-3.5-Sonnet}   & $7.23\pm0.90$ & $3.69\pm0.37$ & $3.27\pm0.14$ & $14.19\pm0.99$ \\
\scalebox{0.95}{\ours}  &	o3-mini	& $7.36\pm0.82$	& $3.73\pm0.25$	& $3.52\pm0.16$	 & $14.61\pm0.84$ \\
\bottomrule
\end{tabular}
\vspace{-5pt}
\caption{The comparative analysis of human evaluation scores, including the calculated mean and standard deviation.}
\label{tab:human_evaluate_results}
\vspace{-10pt}
\end{table*}

The results in Table~\ref{tab:ablation} demonstrate that current settings reach the optimal balanced performance. Although including the visual diagram (\textit{w/} Visual Diagram) shows slightly better performance on the mixed-level metric, upon reviewing the scores of this particular implementation against the default setting, we observe no substantial overall difference. Furthermore, our results demonstrate that the \textit{Paper Lineage} algorithm, coupled with debugging and refinement processes, produces significant improvements across both evaluation metrics.

\subsection{Human Evaluation}
To rigorously assess the reproduction results, we conduct a human evaluation study. For each instance, evaluators are presented with the task instructions, full paper content, reference code, and the generated output. They assess the code across three dimensions: methodological reproducibility, hyperparameter configuration, and the experimental pipeline, assigning maximum scores of 10, 5, and 5, respectively (see Appendix~\ref{sec:huamn_evaluation}). The primary objective of this study is to validate the efficacy of our Mixed-Level metric. Consequently, we calculate and report the Pearson correlation between human judgments and LLM-evaluated scores in Appendix~\ref{sec:huamn_machine_correlation}.
Analysis of the results in Table~\ref{tab:main_results}, \ref{tab:human_evaluate_results}, and \ref{tab:correlation_scores} reveals that mixed-level scores exhibit a stronger correlation with human evaluation than either paper-level or code-level metrics. Qualitative feedback highlights a distinct trend. While LLM agents demonstrate proficiency in reproducing high-level model architectures, their implementations frequently deviate from the reference code regarding granular specifications, such as convolution stride and padding values. Furthermore, the precise reproduction of experimental results is often hindered by the absence of training configurations in source papers, particularly concerning learning rate decay strategies and schedulers.

% \section{Analysis}
\subsection{Paper Lineage Analysis}
\begin{figure}
  \centering
  \includegraphics[width=0.97\linewidth]{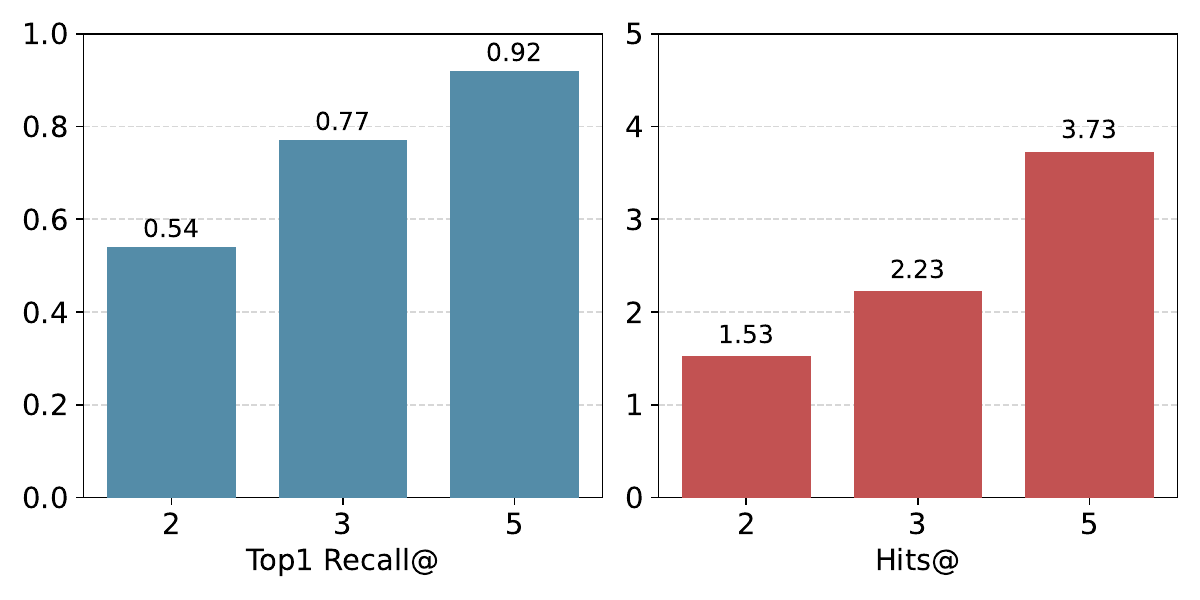}
  \vspace{-10pt}
  \caption{The correlation analysis of papers selected. We utilize \texttt{Claude-3.5-Sonnet} and calculate the mean values of the given metrics.}
  \vspace{-10pt}
  \label{fig:paper_lineage_analysis}
\end{figure}

In our setting, \ours constructs paper lineages by selecting related papers via content and citation analysis. 
To validate the relevance of the papers selected, we establish an expert-curated set of references as standard lineage papers. Specifically, for each source paper, domain experts manually curate five gold-standard references, prioritizing high research relevance and temporal proximity. Search engines such as Google Scholar and Semantic Scholar are utilized to facilitate this selection process.
Given that \ours default selects $k$ papers for each source paper, where $k\in\{2,3,5\}$, we evaluate under two conditions. (i) Mean Top-1 Recall@$k$. Whether the most relevant expert-curated reference is included among the $k$ lineage papers selected by \ours. (ii) Mean Hits@$N$. The number of lineage papers selected by \ours are present within the expert-curated set. The results in Figure~\ref{fig:paper_lineage_analysis} indicate strong agreement between the LLMs and humans in selecting lineage papers.

\subsection{Debugging and Refinement Analysis}
Given that \ours utilizes \texttt{EDIT} command for debugging and refinement, we conduct a further analysis on the details of the code editing. 
To characterize the editing statistics, we define two key metrics: average editing turns and average lines per turn, which are calculated by $N_{\text{turns}}/N_{\text{runs}}$ and $N_{\text{lines}}/N_{\text{turns}}$ respectively. Here, ${N_{\text{lines}}}$ is the total number of edited lines, ${N_{\text{turns }}}$ is the total number of edit turns, and ${N_{\text {runs}}}$ is the total number of successful runs. This analysis exclusively considers executable code generated during the method replication and experiment execution subphases.

The results indicate that utilizing \texttt{o3-mini} and \texttt{Gemini-2.5-Pro} as the LLM backbone are more efficient than \texttt{Claude-3.5-Sonnet}, as evidenced by the fewer debugging and refinement iterations required to converge on high replicability and executable experiment implementations.

\begin{table}[t]
\small
\centering
\setlength{\tabcolsep}{8pt}
\begin{tabular}{l|cc|cc}
\toprule
\multirow{2}{*}{Phase} & \multicolumn{2}{c|}{Debugging} & \multicolumn{2}{c}{Refinement} \\
\cmidrule{2-5} 
& Turns & Lines & Turns & Lines \\
\midrule
\multicolumn{5}{c}{\it{Claude-3.5-Sonnet}} \\
Methods & $5.78$ & $33.52$  & $2.16$ & $40.53$\\
Experiments & $2.84$ & $34.18$ & $1.59$ & $19.29$\\
\midrule
\multicolumn{5}{c}{\it{o3-mini}} \\
Methods & $2.14$ & $21.74$ & $1.23$ & $15.65$ \\
Experiments & $0.94$ & $30.42$ & $0.72$ &  $24.73$\\
\midrule
\multicolumn{5}{c}{\it{Gemini-2.5-Pro}} \\
% Data  &  0.10 & 16.75 & 0.03 & 14 \\
Methods & $1.97$ &  $28.43$ & $1.19$ & $28.10$\\
Experiments & $0.92$ & $26.48$ & $0.78$ & $19.38$\\
\bottomrule
\end{tabular}
\vspace{-5pt}
\caption{The average number of turns and lines when utilizing the \texttt{EDIT} command.}
\label{tab:debugging}
\vspace{-10pt}
\end{table}

\section{Conclusion}
In this study, we explore the automatic experiment reproduction and propose \ours, a multi-agent workflow with paper lineage algorithm and unit testing with a sample batch to generate reproducible and executable code implementations. For evaluation, we introduce \ourbench, an experiment reproduction benchmark containing 13 papers and their human-curated implementation code. This benchmark employs multi-level criteria to assess the performance of generated code, spanning from alignment to execution fidelity. Finally, the result shows that \ours performs superior to other approaches in both \ourbench and PaperBench. We believe that this work will further promote research on paper reproduction and code generation.

\section{Acknowledgment}
The work was initiated and supported by the AI9Stars Team. We are grateful for the support of the OpenBMB team.

\section*{Limitations}
In this work, we investigate the automatic reproducibility of AI experiments, a critical aspect we believe will significantly advance automation within the AI field. However, our current approach has certain limitations that present avenues for future research. Primarily, our method is specialized for replicating individual experimental tasks rather than performing broader code generation at the repository level. Consequently, enhancing execution capabilities within the context of repo-level code warrants further exploration. Furthermore, the inherent complexity of raw datasets often necessitates preliminary data processing. Automating this data preprocessing stage represents a substantial research challenge that must be addressed to achieve more comprehensive automation.

\section*{Ethics Statement}
The primary objective of this work is to automate the replication of experiments detailed in existing, publicly available research papers. While the methodologies are drawn from the public domain, which generally implies transparency, it is important to acknowledge the potential for data leakage associated with using our system. Users should therefore be mindful of this possibility, particularly when the replication process might involve sensitive datasets or generate intermediate results that could inadvertently disclose information. We provide a license check in the Appendix for the License of papers and code in the \ourbench. Furthermore, LLMs are utilized for writing and refining the paper contents.

\bibliography{custom}

@article{gu2024large,
  title={Large language models for constructing and optimizing machine learning workflows: A survey},
  author={Gu, Yang and You, Hengyu and Cao, Jian and Yu, Muran and Fan, Haoran and Qian, Shiyou},
  journal={arXiv preprint arXiv:2411.10478},
  year={2024}
}

@article{zhang2023automl,
  title={Automl-gpt: Automatic machine learning with gpt},
  author={Zhang, Shujian and Gong, Chengyue and Wu, Lemeng and Liu, Xingchao and Zhou, Mingyuan},
  journal={arXiv preprint arXiv:2305.02499},
  year={2023}
}

@article{liu2023jarvix,
  title={JarviX: A LLM no code platform for tabular data analysis and optimization},
  author={Liu, Shang-Ching and Wang, ShengKun and Lin, Wenqi and Hsiung, Chung-Wei and Hsieh, Yi-Chen and Cheng, Yu-Ping and Luo, Sian-Hong and Chang, Tsungyao and Zhang, Jianwei},
  journal={arXiv preprint arXiv:2312.02213},
  year={2023}
}

@article{jeong2024llm,
  title={Llm-select: Feature selection with large language models},
  author={Jeong, Daniel P and Lipton, Zachary C and Ravikumar, Pradeep},
  journal={arXiv preprint arXiv:2407.02694},
  year={2024}
}

@inproceedings{yang2025autommlab,
  title={Autommlab: Automatically generating deployable models from language instructions for computer vision tasks},
  author={Yang, Zekang and Zeng, Wang and Jin, Sheng and Qian, Chen and Luo, Ping and Liu, Wentao},
  booktitle={Proceedings of the AAAI Conference on Artificial Intelligence},
  volume={39},
  number={21},
  pages={22056--22064},
  year={2025}
}

@inproceedings{luo2024autom3l,
  title={Autom3l: An automated multimodal machine learning framework with large language models},
  author={Luo, Daqin and Feng, Chengjian and Nong, Yuxuan and Shen, Yiqing},
  booktitle={Proceedings of the 32nd ACM International Conference on Multimedia},
  pages={8586--8594},
  year={2024}
}

@article{tang2024modelgpt,
  title={Modelgpt: Unleashing llm's capabilities for tailored model generation},
  author={Tang, Zihao and Lv, Zheqi and Zhang, Shengyu and Wu, Fei and Kuang, Kun},
  journal={arXiv preprint arXiv:2402.12408},
  year={2024}
}

@article{hollmann2022tabpfn,
  title={Tabpfn: A transformer that solves small tabular classification problems in a second},
  author={Hollmann, Noah and M{\"u}ller, Samuel and Eggensperger, Katharina and Hutter, Frank},
  journal={arXiv preprint arXiv:2207.01848},
  year={2022}
}

@article{erdil2023explosive,
  title={Explosive growth from AI automation: A review of the arguments},
  author={Erdil, Ege and Besiroglu, Tamay},
  journal={arXiv preprint arXiv:2309.11690},
  year={2023}
}

@article{weng2024cycleresearcher,
  title={Cycleresearcher: Improving automated research via automated review},
  author={Weng, Yixuan and Zhu, Minjun and Bao, Guangsheng and Zhang, Hongbo and Wang, Jindong and Zhang, Yue and Yang, Linyi},
  journal={arXiv preprint arXiv:2411.00816},
  year={2024}
}

@inproceedings{wang2024scimon,
  title={Scimon: Scientific inspiration machines optimized for novelty},
  author={Wang, Qingyun and Downey, Doug and Ji, Heng and Hope, Tom},
  booktitle={Proceedings of the 62nd Annual Meeting of the Association for Computational Linguistics (Volume 1: Long Papers)},
  pages={279--299},
  year={2024}
}

@article{schmidgall2025agentrxiv,
  title={Agentrxiv: Towards collaborative autonomous research},
  author={Schmidgall, Samuel and Moor, Michael},
  journal={arXiv preprint arXiv:2503.18102},
  year={2025}
}

@article{qian2023chatdev,
  title={Chatdev: Communicative agents for software development},
  author={Qian, Chen and Liu, Wei and Liu, Hongzhang and Chen, Nuo and Dang, Yufan and Li, Jiahao and Yang, Cheng and Chen, Weize and Su, Yusheng and Cong, Xin and others},
  journal={arXiv preprint arXiv:2307.07924},
  year={2023}
}

@article{bogin2024super,
  title={Super: Evaluating agents on setting up and executing tasks from research repositories},
  author={Bogin, Ben and Yang, Kejuan and Gupta, Shashank and Richardson, Kyle and Bransom, Erin and Clark, Peter and Sabharwal, Ashish and Khot, Tushar},
  journal={arXiv preprint arXiv:2409.07440},
  year={2024}
}

@article{gandhi2025researchcodeagent,
  title={ResearchCodeAgent: An LLM Multi-Agent System for Automated Codification of Research Methodologies},
  author={Gandhi, Shubham and Shah, Dhruv and Patwardhan, Manasi and Vig, Lovekesh and Shroff, Gautam},
  journal={arXiv preprint arXiv:2504.20117},
  year={2025}
}

@article{seo2025paper2code,
  title={Paper2Code: Automating Code Generation from Scientific Papers in Machine Learning},
  author={Seo, Minju and Baek, Jinheon and Lee, Seongyun and Hwang, Sung Ju},
  journal={arXiv preprint arXiv:2504.17192},
  year={2025}
}

@article{xi2025rise,
  title={The rise and potential of large language model based agents: A survey},
  author={Xi, Zhiheng and Chen, Wenxiang and Guo, Xin and He, Wei and Ding, Yiwen and Hong, Boyang and Zhang, Ming and Wang, Junzhe and Jin, Senjie and Zhou, Enyu and others},
  journal={Science China Information Sciences},
  volume={68},
  number={2},
  pages={121101},
  year={2025},
  publisher={Springer}
}

@article{starace2025paperbench,
  title={PaperBench: Evaluating AI's Ability to Replicate AI Research},
  author={Starace, Giulio and Jaffe, Oliver and Sherburn, Dane and Aung, James and Chan, Jun Shern and Maksin, Leon and Dias, Rachel and Mays, Evan and Kinsella, Benjamin and Thompson, Wyatt and others},
  journal={arXiv preprint arXiv:2504.01848},
  year={2025}
}

@article{siegel2024core,
  title={CORE-Bench: Fostering the Credibility of Published Research Through a Computational Reproducibility Agent Benchmark},
  author={Siegel, Zachary S and Kapoor, Sayash and Nagdir, Nitya and Stroebl, Benedikt and Narayanan, Arvind},
  journal={arXiv preprint arXiv:2409.11363},
  year={2024}
}

@article{si2024can,
  title={Can llms generate novel research ideas? a large-scale human study with 100+ nlp researchers},
  author={Si, Chenglei and Yang, Diyi and Hashimoto, Tatsunori},
  journal={arXiv preprint arXiv:2409.04109},
  year={2024}
}

@article{li2024chain,
  title={Chain of ideas: Revolutionizing research via novel idea development with llm agents},
  author={Li, Long and Xu, Weiwen and Guo, Jiayan and Zhao, Ruochen and Li, Xingxuan and Yuan, Yuqian and Zhang, Boqiang and Jiang, Yuming and Xin, Yifei and Dang, Ronghao and others},
  journal={arXiv preprint arXiv:2410.13185},
  year={2024}
}

@article{shao2023iebins,
  title={Iebins: Iterative elastic bins for monocular depth estimation},
  author={Shao, Shuwei and Pei, Zhongcai and Wu, Xingming and Liu, Zhong and Chen, Weihai and Li, Zhengguo},
  journal={Advances in Neural Information Processing Systems},
  volume={36},
  pages={53025--53037},
  year={2023}
}

@article{liu2023itransformer,
  title={itransformer: Inverted transformers are effective for time series forecasting},
  author={Liu, Yong and Hu, Tengge and Zhang, Haoran and Wu, Haixu and Wang, Shiyu and Ma, Lintao and Long, Mingsheng},
  journal={arXiv preprint arXiv:2310.06625},
  year={2023}
}

@inproceedings{zhao2022decoupled,
  title={Decoupled knowledge distillation},
  author={Zhao, Borui and Cui, Quan and Song, Renjie and Qiu, Yiyu and Liang, Jiajun},
  booktitle={Proceedings of the IEEE/CVF Conference on computer vision and pattern recognition},
  pages={11953--11962},
  year={2022}
}

@inproceedings{gao2022simvp,
  title={Simvp: Simpler yet better video prediction},
  author={Gao, Zhangyang and Tan, Cheng and Wu, Lirong and Li, Stan Z},
  booktitle={Proceedings of the IEEE/CVF conference on computer vision and pattern recognition},
  pages={3170--3180},
  year={2022}
}

@inproceedings{chen2023humanmac,
  title={Humanmac: Masked motion completion for human motion prediction},
  author={Chen, Ling-Hao and Zhang, Jiawei and Li, Yewen and Pang, Yiren and Xia, Xiaobo and Liu, Tongliang},
  booktitle={Proceedings of the IEEE/CVF international conference on computer vision},
  pages={9544--9555},
  year={2023}
}

@inproceedings{cui2023selective,
  title={Selective frequency network for image restoration},
  author={Cui, Yuning and Tao, Yi and Bing, Zhenshan and Ren, Wenqi and Gao, Xinwei and Cao, Xiaochun and Huang, Kai and Knoll, Alois},
  booktitle={The eleventh international conference on learning representations},
  year={2023}
}

@article{wu2023solving,
  title={Solving high-dimensional pdes with latent spectral models},
  author={Wu, Haixu and Hu, Tengge and Luo, Huakun and Wang, Jianmin and Long, Mingsheng},
  journal={arXiv preprint arXiv:2301.12664},
  year={2023}
}

@inproceedings{cao2022swin,
  title={Swin-unet: Unet-like pure transformer for medical image segmentation},
  author={Cao, Hu and Wang, Yueyue and Chen, Joy and Jiang, Dongsheng and Zhang, Xiaopeng and Tian, Qi and Wang, Manning},
  booktitle={European conference on computer vision},
  pages={205--218},
  year={2022},
  organization={Springer}
}

@inproceedings{wang2021tree,
  title={Tree decomposed graph neural network},
  author={Wang, Yu and Derr, Tyler},
  booktitle={Proceedings of the 30th ACM international conference on information \& knowledge management},
  pages={2040--2049},
  year={2021}
}

@article{desai2021timevae,
  title={Timevae: A variational auto-encoder for multivariate time series generation},
  author={Desai, Abhyuday and Freeman, Cynthia and Wang, Zuhui and Beaver, Ian},
  journal={arXiv preprint arXiv:2111.08095},
  year={2021}
}

@article{jiang2023low,
  title={Low-light image enhancement with wavelet-based diffusion models},
  author={Jiang, Hai and Luo, Ao and Fan, Haoqiang and Han, Songchen and Liu, Shuaicheng},
  journal={ACM Transactions on Graphics (TOG)},
  volume={42},
  number={6},
  pages={1--14},
  year={2023},
  publisher={ACM New York, NY, USA}
}

@inproceedings{choi2023blurring,
  title={Blurring-sharpening process models for collaborative filtering},
  author={Choi, Jeongwhan and Hong, Seoyoung and Park, Noseong and Cho, Sung-Bae},
  booktitle={Proceedings of the 46th international ACM SIGIR conference on research and development in information retrieval},
  pages={1096--1106},
  year={2023}
}

@inproceedings{chen2023dual,
  title={Dual aggregation transformer for image super-resolution},
  author={Chen, Zheng and Zhang, Yulun and Gu, Jinjin and Kong, Linghe and Yang, Xiaokang and Yu, Fisher},
  booktitle={Proceedings of the IEEE/CVF international conference on computer vision},
  pages={12312--12321},
  year={2023}
}

@article{schmidgall2025agent,
  title={Agent laboratory: Using llm agents as research assistants},
  author={Schmidgall, Samuel and Su, Yusheng and Wang, Ze and Sun, Ximeng and Wu, Jialian and Yu, Xiaodong and Liu, Jiang and Liu, Zicheng and Barsoum, Emad},
  journal={arXiv preprint arXiv:2501.04227},
  year={2025}
}

@article{wang2024mineru,
  title={Mineru: An open-source solution for precise document content extraction},
  author={Wang, Bin and Xu, Chao and Zhao, Xiaomeng and Ouyang, Linke and Wu, Fan and Zhao, Zhiyuan and Xu, Rui and Liu, Kaiwen and Qu, Yuan and Shang, Fukai and others},
  journal={arXiv preprint arXiv:2409.18839},
  year={2024}
}

@article{baek2024researchagent,
  title={Researchagent: Iterative research idea generation over scientific literature with large language models},
  author={Baek, Jinheon and Jauhar, Sujay Kumar and Cucerzan, Silviu and Hwang, Sung Ju},
  journal={arXiv preprint arXiv:2404.07738},
  year={2024}
}

@inproceedings{wang2024executable,
  title={Executable code actions elicit better llm agents},
  author={Wang, Xingyao and Chen, Yangyi and Yuan, Lifan and Zhang, Yizhe and Li, Yunzhu and Peng, Hao and Ji, Heng},
  booktitle={Forty-first International Conference on Machine Learning},
  year={2024}
}

@article{nie2022time,
  title={A time series is worth 64 words: Long-term forecasting with transformers},
  author={Nie, Yuqi and Nguyen, Nam H and Sinthong, Phanwadee and Kalagnanam, Jayant},
  journal={arXiv preprint arXiv:2211.14730},
  year={2022}
}

@article{chen2024transunet,
  title={TransUNet: Rethinking the U-Net architecture design for medical image segmentation through the lens of transformers},
  author={Chen, Jieneng and Mei, Jieru and Li, Xianhang and Lu, Yongyi and Yu, Qihang and Wei, Qingyue and Luo, Xiangde and Xie, Yutong and Adeli, Ehsan and Wang, Yan and others},
  journal={Medical Image Analysis},
  volume={97},
  pages={103280},
  year={2024},
  publisher={Elsevier}
}

@article{li2025system,
  title={From system 1 to system 2: A survey of reasoning large language models},
  author={Li, Zhong-Zhi and Zhang, Duzhen and Zhang, Ming-Liang and Zhang, Jiaxin and Liu, Zengyan and Yao, Yuxuan and Xu, Haotian and Zheng, Junhao and Wang, Pei-Jie and Chen, Xiuyi and others},
  journal={arXiv preprint arXiv:2502.17419},
  year={2025}
}

@misc{lin2025autop2cllmbasedagentframework,
      title={AutoP2C: An LLM-Based Agent Framework for Code Repository Generation from Multimodal Content in Academic Papers}, 
      author={Zijie Lin and Yiqing Shen and Qilin Cai and He Sun and Jinrui Zhou and Mingjun Xiao},
      year={2025},
      eprint={2504.20115},
      archivePrefix={arXiv},
      primaryClass={cs.SE},
}
\clearpage
\appendix

\section{Appendix}
\label{sec:appendix}

\subsection{Execution Error Analysis}
A manual error analysis of code generated by both PaperCoder \cite{seo2025paper2code} and \ours reveals that the majority of issues stemmed from incorrect data shapes during internal model calculations. Intuitively, while LLMs can generate specified network structures, achieving correct data flow without test data is challenging, particularly for complex architectures. 
For tasks involving pre-trained models, it is often not possible to generate a completely correct model architecture implementation in a single attempts. This is mainly because LLMs still face minor challenges in accurately reproducing architectural implementations of common pre-trained models, for example, hidden layer dimensions and normalization layer configurations.
Although debugging resolves numerous errors, some issues persist, particularly when encountering complex data flows.
% which are computed through various modules.

\subsection{\ours Details}
For all the experiments conducted in the experiment section, \ours utilizes \texttt{claude-3-5-sonnet-20240620} version as \texttt{claude-3-5-sonnet} backbone.
In the code development phase of \ours, the code agent debugs the execution errors. We set the maximum debug tries to 20 for all the subphases in the code development phase. However, further debug attempts generally do not produce more executable code, as bugs are typically fixed within 5-8 iterations. The results are also denoted in Table~\ref{tab:debugging}. On the \ourbench benchmark, the average cost for \ours to reproduce a single experiment is \$$1.87$ when utilising the \texttt{o3-mini} as the LLM backbone. We execuate all the generated code on the Tesla A100 GPUs.

The algorithm for paper lineage is provided in the Algorithm \ref{alg:paper_lineage}.

\subsection{\ourbench Details}

While our main text provides an overview of the research domain, datasets, and evaluation metrics integral to the \ourbench, this section offers a more detailed exposition of these elements.
\subsubsection{Reference Code}
We count the total lines of code for this reference implementation and its associated preprocessing code. Also, statistics about whether pretrained models are loaded and the reference performance rerun by ourselves are conducted. The details are shown in Table~\ref{tab:benchmark_appendix}. Due to potential differences between our rerun settings and the experimental setup originally used by the paper authors, the reference performance may differ from the performance metric reported in their respective papers. The reference performances that we utilize in calculating performance gap metrics are obtained by rerunning the code on Tesla A100 GPUs.

\begin{table*}[h]
\centering
\setlength{\tabcolsep}{8pt}
\begin{tabular}{lcccccc}
\toprule
\multirow{2}{*}{\textbf{Method}} & \textbf{Total} &\textbf{Preprocess} & \textbf{Preprocess} & \textbf{Pretrained} & \textbf{Evaluation} & \textbf{Reference} \\
 & \textbf{Lines}  &  \textbf{Code} & \textbf{Lines} &\textbf{Model} & \textbf{Metric}   & \textbf{Performance} \\
\midrule
\emph{IEBins} & 1877 & \cmark & 320 & \xmark & $\delta < 1.25$ & $0.8854$\\
\emph{iTransformer} & 610 &  \cmark & 199 & \xmark & MSE & $0.4008$\\    
\emph{DKD} & 678 & \xmark & 0 & \cmark & Accuracy & $74.56$\\
\emph{SimVP} & 544 &  \cmark & 161 & \xmark & MSE & $26.19$\\
\emph{HumanMAC}  & 1196 & \cmark & 412 & \xmark & ADE  &$0.2195$\\
\emph{SFNet} & 805 & \cmark & 121 & \xmark & PSNR & $39.68$\\
\emph{LSM} & 403 & \cmark & 141 & \xmark & MSE & $0.0074$\\
\emph{Swin-Unet} & 1218 & \cmark & 102 & \cmark & DSC & $0.7309$\\
\emph{TDGNN-w} & 310 & \cmark & 21 & \xmark & Accuracy & $0.7651$\\
% \emph{NBFNet}~\citep{zhu2021neural} & Link Prediction & WN18RR & H@10\\
\emph{TimeVAE} & 711 & \cmark  & 69 & \xmark & Predictor & $0.2083$ \\
\emph{WCDM} & 935 & \cmark & 117 &\xmark & SSIM & $0.7938$\\
\emph{BSPM} & 713 & \cmark & 221 & \xmark & Recall & $0.1921$\\
\emph{DAT-S} & 2047 &  \cmark  & 163 & \xmark  & PSNR & $38.48$\\
\bottomrule
\end{tabular}
\vspace{-5pt}
\caption{More details about \ourbench. The Reference Performance is the rerun performance obtained by official implementations. We utilize them to calculate the performance gap.}
\label{tab:benchmark_appendix}
\end{table*}

\begin{algorithm}[t]
\caption{Paper Lineage Algorithm}
\label{alg:paper_lineage}
\begin{algorithmic}[1]
\State \textbf{Input:} Paper $\mathcal{P}$; research agent $\mathcal{RA}$; code agent $\mathcal{CA}$; Integer $k$ (default $3$); Instructions $\mathcal{I}$.
\State \textbf{Output:} Set of paper lineage elements $K_{\text{lineage}}$.

\State Initialize $K_{\text{lineage}}$.

\State $\{P_1,...P_k\} \gets \mathcal{RA}(\mathcal{P}, k)$. \Comment{research agent identifies top-$k$ relevant papers}

\For{each paper $P_i$ in $\{P_i,...P_k\}$}
    \State $\text{Text}_i \gets \operatorname{DownloadPaper}(P_i)$. \Comment{e.g., via ArXiv/Semantic Scholar API}
    \State $(\mathcal{S}_i, \mathcal{U}_i) \gets \mathcal{RA}(\text{Text}_i)$. \Comment{research agent extracts summary \& repo URL}
    
    \State $\mathcal{K}_i \gets S_i$. \Comment{Initialize $\mathcal{K}_i$ with summary}
    \If{$U_i \text{ is valid and accessible}$}
        \State $\text{Repo}_i \gets \operatorname{DownloadRepo}(U_i)$. \Comment{e.g., via GitHub API}
        \State $\text{Code}_i \gets \mathcal{CA}(S_i, \text{Repo}_i, \mathcal{I})$. \Comment{code agent filters for relevant code}
        \State $\mathcal{K}_i \gets \langle S_i, \text{Code}_i \rangle$. \Comment{Update $\mathcal{K}_i$ to summary-code pair}
    \EndIf
    \State Add $\mathcal{K}_i$ to $K_{\text{lineage}}$.
\EndFor
\State \textbf{Return} $K_{\text{lineage}}$.
\end{algorithmic}
\end{algorithm}

\subsubsection{Automatic Evaluation}
For the evaluation metric in the align-score, we utilize LLMs to judge the generated code from three aspects, including paper-level, code-level and mixed-level. The prompts provided to the judge LLM are generally long, requiring a powerful model for robust evaluations. Consequently, we employ \texttt{o1} for this purpose. For each level, each reproduction requires approximately 
\$$0.5$ for evaluation. Thus, an evaluation run to determine the align-score costs approximately \$$20$ on \ourbench. When evaluating the generated code in the repository \cite{qian2023chatdev}, all the Python files are concatenated to form the generated code. Figure~\ref{fig:prompt_summarize} denotes the prompt to extract 5 key points from the paper by utilizing \texttt{o1}. Figure~\ref{fig:code_high_level} shows the prompt for paper-level evaluation and Figure~\ref{fig:code_low_level1}~\ref{fig:code_low_level2} show the prompt for code-level evaluation.  Figure~\ref{fig:code_multi_level1}~\ref{fig:code_multi_level2} shows the prompt for mixed-level evaluation. The placeholder in each prompt is replaced with the corresponding content.

\subsection{Human Evaluation Details}\label{sec:huamn_evaluation}
\subsubsection{Evaluation Settings}
For the human evaluations, the evaluators are given the original papers, instructions, official implementations, and generated ones. To reduce the workload of human assessment, we simultaneously generated summaries of the papers using LLMs, which helps to accelerate the evaluators' understanding of the papers. We recruit 5 students for human evaluations, including 3 PhDs, 1 master and a senior undergraduate. All the evaluators are from the EECS-related majors. The overall instructions are like the prompts for mixed-level evaluation denoted in Figure \ref{human_eval_instruction1}, \ref{human_eval_instruction2}. The human evaluators should score the generated code through the completeness of the method, parameters and experiment pipeline with a maximum of 10, 5, and 5. The score criteria are similar to the mixed-level evaluations in Figure~\ref{fig:code_multi_level1} and \ref{fig:code_multi_level2}. The minimum and maximum scores only occur in extreme circumstances where the code is totally wrong or exactly correct. For each implementation, the evaluators need around 15-25 minutes for the evaluation. 

\subsubsection{Human-Machine Score correlation}\label{sec:huamn_machine_correlation}
To evaluate the correlation of the alignment score evaluated by the LLM judge and the human evaluation scores, we calculate the Pearson correlation coefficients of these two scores. We evaluate all three runs in the main experiments. Since there is no one-to-one correspondence between the above two scores, we calculate the correlation by comparing the three alignment scores against the single average human score for each task. The results in Table~\ref{tab:correlation_scores} demonstrate that the mixed-level score aligns the best with the human scores, which proves our proposed evaluation strategy.

\begin{table*}[h]
\centering
\begin{tabular}{lcccc}
\toprule
\textbf{Method} & \textbf{LLM} & \textbf{Paper-level} & \textbf{Code-level} & \textbf{Mixed-level} \\ 
\midrule
\ours   & o3-mini      & 0.48   &0.76  &0.81 \\
PaperCoder      & o3-mini      & 0.52   &0.73  &0.83 \\
\midrule
\ours   & Claude-3.5-Sonnet   & 0.49   &0.71  &0.78 \\
ChatDev         & GPT-4o       & 0.37   &0.79  &0.81 \\ 
\bottomrule
\end{tabular}
\caption{The human-machine score correlation between the human and alignment scores.}
\label{tab:correlation_scores}
\end{table*}

\subsection{License Check}
\ours is a non-profit project intended solely for research purposes. In constructing the \ourbench, we carefully curate the datasets, selecting only those licenses for research and downloading them in strict accordance with any specified requirements. For the paper and code, we have re-examined the ground-truth code for all papers and found that it is all under the MIT and Apache-2.0 licenses. The specific terms state that: 
Permission is hereby granted, free of charge, to any person obtaining a copy of this software and associated documentation files (the 'Software'), to deal in the Software without restriction, including without limitation the rights to use, copy, modify, merge, publish, distribute, sublicense, and/or sell copies of the Software. Secondly, we have screened the papers themselves, and their licenses are CC BY-NC-ND (Attribution-NonCommercial-NoDerivatives International), which denotes they are free to: Share — copy and redistribute the material in any medium or format. Due to our non-profit nature, we comply with all the aforementioned terms. Additionally, we analyze the lineage papers and code for \ours. Since the lineage papers are obtained via the ArXiv API, the permissions for papers on ArXiv fall under the following licenses: 
\begin{itemize}[leftmargin=*]
    \item  CC BY: Creative Commons Attribution
    \item CC BY-SA: Creative Commons Attribution-ShareAlike
    \item CC BY-NC-SA: Creative Commons Attribution-Noncommercial-ShareAlike
    \item CC BY-NC-ND: Creative Commons Attribution-NonCommercial-NoDerivatives
    \item CC Zero: No Rights Reserved
\end{itemize}

Therefore, we comply with all the usage regulations given our non-commercial purpose. Furthermore, we have checked the corresponding code repositories and found that they all fall under the MIT License, Apache-2.0 licenses, and the BSD License. As our work only involves refactoring this code, our actions comply with its relevant requirements.

\begin{figure*}[ht]
\begin{tcolorbox}[colback=white, colframe=black, title=Input queries of \ours]
\small\texttt{ARXIV\_ID='\textcolor{blue}{ArXiv ID}'\\
TASK='\textcolor{blue}{Task Name}'\\
MODEL='\textcolor{blue}{Model Name}'\\
METRIC='\textcolor{blue}{Evaluation Metric}'\\
INSTRUCTION=f"""You are assigned an arXiv paper to replicate.
You need to replicate the experiment conducted for \{TASK\} dataset. Utilizing \{METRIC\} as the evaluation metric. The dataset/checkpoints is under '\textcolor{blue}{Relative Path}'}
\end{tcolorbox}
\vspace{-5pt}
\caption{The input quries of \ours. It contains the ARXIV ID to download the paper. TASK, MODEL and METRIC in the paper that need to be reproduced.}
\label{fig:input_queries_prompt}
\end{figure*}   

\begin{figure*}[ht]
\begin{tcolorbox}[colback=white, colframe=black, title=Iterative dialogue template of LLM agents]
\small\texttt{\textasciitilde\textasciitilde\textasciitilde\textasciitilde\textasciitilde\textasciitilde\textasciitilde\textasciitilde\textasciitilde \\
History: \{\textcolor{blue}{history string}\}\\
\textasciitilde\textasciitilde\textasciitilde\textasciitilde\textasciitilde\textasciitilde\textasciitilde\textasciitilde\textasciitilde \\
Current Step: \{\textcolor{blue}{step}\}, Phase: \{\textcolor{blue}{phase}\} \\
Task instructions: \{\textcolor{blue}{current phase prompts}\} \\
{}[Overall Objective] Your overall goal is to follow the instructions to replicate the method proposed in the paper. \\
Instruction: \{\textcolor{blue}{instruction}\}. \\
To achieve the objective, start by conducting a literature review, learning relevant codes, and finally generating the method and experiment codes. \\
\{\textcolor{blue}{previous step feedback}\}\\
\{\textcolor{blue}{command instruction}\} \\
\{\textcolor{blue}{additional notes}\} \\
When you are given commands that you can use, your reply must be selected from among the commands. Please produce a single command below: }
\end{tcolorbox}
\caption{Prompt for the \texttt{ADD} command of code agent.}
\label{fig:dialogue_template_prompt}
\end{figure*}

\begin{figure*}[ht]
\begin{tcolorbox}[colback=white, colframe=black, title=Abbreviation prompts for Paper Lineage stage of the research agent]
\small\texttt{Your task is to read the paper and identify the 3 most relevant papers from its references that help in understanding the paper's contributions, including the proposed model architecture, experimental settings, and other details. These papers need to be in the same research field as the ones that need to be replicated. \\
You need to infer the most relevant related works based on the information such as the position and name of the reference. Importantly, the name of the paper should be correct. Do not generate mismatched names. \\
The selected papers must come from the references and be specific to the same research field as the paper, avoiding commonly cited works like 'Attention Is All You Need'. Also, do not add the paper in the instructions. Return the related works in the format: ['paper name 1', 'paper name 2', ...] with only paper names (author names should not be included). \\
Add the most related work after reading using the following command: \\
\textasciigrave\textasciigrave\textasciigrave ADD\textbackslash n<related code file>\textbackslash n\textasciigrave\textasciigrave\textasciigrave \\
where <related work list> is the list of related work in the reference, the item in the list should be the full name of the paper. ADD is just the word ADD. \\
You can only use a single command per inference turn. Do not use more than one command per inference. If you use multiple commands, then only one of them will be executed, not both.\\
}
\end{tcolorbox}
\vspace{-5pt}
\caption{Abbreviation prompts for Paper Lineage stage of the research agent}
\label{fig:paper_lineage_prompt_research_agent}
\end{figure*}        

\begin{figure*}[ht]
\begin{tcolorbox}[colback=white, colframe=black, title=Abbreviation prompts for Paper Lineage stage of the code agent]
\small\texttt{You need to reproduce the \{\textcolor{blue}{Task Instruction}\} experiment now. The code repository mentioned above is related to the target reproduction paper. Please filter out the code files that are helpful for the reproduction based on the instructions, and return the useful code files in the form of a list. For example, \textasciigrave\textasciigrave\textasciigrave python ['file1.py', 'model/file2.py']\textasciigrave\textasciigrave\textasciigrave. You should start from the most relevant code file. \\
You could select related code files using the following command: \\
\textasciigrave\textasciigrave\textasciigrave ADD\textbackslash n<related code file>\textbackslash n\textasciigrave\textasciigrave\textasciigrave \\
where <related code file> is the list of related code files in the given code repo, the item in the list should be the full name of the file path. ADD is just the word ADD.
}
\end{tcolorbox}
\vspace{-5pt}
\caption{Abbreviation prompts for Paper Lineage stage of the code agent}
\label{fig:paper_lineage_prompt_code_agent}
\end{figure*}    
                
\begin{figure*}[ht]
\begin{tcolorbox}[colback=white, colframe=black, title=Code agent prompt for the \texttt{EDIT} command.]
\small\texttt{You can edit code using the following command: \\
\textcolor{blue}{\textasciigrave\textasciigrave\textasciigrave EDIT\textbackslash n N M\textbackslash n<new code>\textbackslash n\textasciigrave\textasciigrave\textasciigrave} \\
EDIT is the word EDIT, N is the first line index you want to replace, and M is the last line index you want to replace (everything in between will also be removed), and <new code> will be the new code that is replacing the old code.\\
This command allows you to replace lines indexed n through m (n:m) of the current code with as many lines of new code as you want to add. This will be the primary way that you interact with code. \\   You can only use a single command per inference turn. Do not use more than one command per inference. If you use multiple commands, then only one of them will be executed, not both.}
\end{tcolorbox}
\vspace{-5pt}
\caption{Prompt for the \texttt{EDIT} command of code agent.}
\label{fig:edit_prompt}
\end{figure*}

\begin{figure*}[ht]
\begin{tcolorbox}[colback=white, colframe=black, title=Abbreviation of human evaluation instructions.]
\small\texttt{
Standards for manual score evaluation \\
1. Overview \& Objective\\
You are acting as an expert evaluator to assess the quality and fidelity of LLM-generated code. Your task is to compare \textcolor{blue}{Generated Code} against the official \textcolor{blue}{Reference Code} (Ground Truth) for a specific research paper.\\
Goal: Determine how accurately the generated code reproduces the specific methods, parameters, and experimental pipeline described in the paper and implemented in the reference code.\\
2. Scoring Criteria (Total: 20 Points)\\
\\
Please evaluate the code across three specific dimensions. Use the \textcolor{blue}{Reference Code} as the absolute standard for correctness.\\
\\
A. Completeness of Method (Max 10 Points)\\
Focus: Does the code implement the core modeling innovation, specific algorithms, network architecture, and loss functions?\\
0 - 1 Points (Total Difference):\\
The core innovation is missing or completely incorrect. The code might implement a standard baseline (e.g., standard ResNet) instead of the paper's proposed method, or the logic is fundamentally flawed and unrunnable.\\
2 - 4 Points (Unsimilar):\\
The code attempts the method but misses critical components. Major modules are absent, or the mathematical logic (e.g., attention mechanism details, specific loss calculation) deviates significantly from the reference.\\
5 - 7 Points (Similar):\\
The core concepts are present. The implementation captures the "essence" of the method, but there are noticeable inaccuracies, over-simplifications, or structural differences compared to the reference code.\\
8 - 9 Points (Roughly the Same):\\
High fidelity. The logic and structure closely mirror the reference code. Differences are superficial (e.g., variable naming, modularization style) and do not affect the core functionality.\\
10 Points (Same):
Functionally identical. The implementation of the key method is a near-perfect replica of the reference code logic.\\
\\
B. Parameters (Max 5 Points)\\
Focus: Does the code use the correct hyperparameters, dimensions, and constants as specified in the reference code?\\
0 Points (Total Mismatch):
Uses generic library defaults (e.g., `lr=1e-3`, `hidden\_dim=256`) or random values that completely contradict the paper’s specific setup.\\
1 Point (Significant Deviation):\\
Attempts to configure parameters, but critical architectural dimensions (e.g., number of layers, embedding sizes, distinct model depths) are incorrect relative to the reference.\\
2 Points (Partial Match):\\
The model architecture parameters are mostly correct, but critical training hyperparameters (e.g., specific learning rate schedules, weight decay, optimizer betas) or loss coefficients are missing or wrong.\\
3 Points (Moderate Match):\\
The majority of parameters (both model and training) align with the reference, but there are noticeable discrepancies in specific configuration details (e.g., wrong kernel size in a specific layer, incorrect temperature scaling value).\\
4 Points (High Fidelity):\\
The configuration is nearly identical to the reference code. The discrepancies are negligible and non-critical (e.g., a slightly different buffer size or a different variable name for a constant) that do not impact the main results.\\
5 Points (Exact Match):\\
Perfect replication. The generated code strictly adheres to all hyperparameters, model dimensions, and configuration settings found in the reference code without any error.
}
\end{tcolorbox}
\vspace{-5pt}
\caption{Prompt for human evaluation instructions.}
\label{fig:human_eval_instruction1}
\end{figure*}

\begin{figure*}[ht]
\begin{tcolorbox}[colback=white, colframe=black, title=Abbreviation of human evaluation instructions.]
\small\texttt{
C. Experiment Pipeline (Max 5 Points)\\
Focus: Is the data processing, training loop, and evaluation protocol correct?\\
0 Points (Broken / Wrong Task):\\
The pipeline is non-functional, syntax-heavy errors prevent execution, or it solves a completely different task (e.g., generating a classification loop for a segmentation paper).\\
1 Point (Incorrect Data Handling):\\
The training loop structure exists, but the data loading or input formatting is fundamentally wrong or hallucinated (e.g., using a fake dataset class that doesn't exist or handling data dimensions incorrectly).\\
2 Points (Generic / Simplified): \\
The code provides a standard "boilerplate" pipeline (standard loader $\to$ model $\to$ loss). It works, but misses all paper-specific customizations like specific data augmentations, sampling strategies, or preprocessing steps. \\
3 Points (Incomplete Protocol): \\
The pipeline flow is correct and includes some specific steps, but misses critical evaluation metrics or specific evaluation procedures defined in the paper (e.g., using simple Accuracy instead of a specific mIoU calculation, or missing a required post-processing step). \\
4 Points (High Consistency):\\
The pipeline is robust and follows the correct protocol (Data Load $\to$ Model $\to$ Loss $\to$ Metric). The logic matches the reference, with only minor deviations in non-critical areas like logging, random seed setting, or exact validation split ratios.\\
5 Points (Exact Replication):\\
Perfect reproduction. The code includes specific preprocessing logic, exact data splits, specific augmentation pipelines, and metric calculations exactly as implemented in the reference code.\\
\\
3. Evaluation Guidelines\\
1.  Logic over Comments: Ignore comments in the code. Judge based on the executable logic/code statements.\\
2.  Functional Equivalence: If the generated code achieves the same mathematical result as the reference but uses a slightly different coding style (e.g., 2 lines vs 1 line), consider it correct.\\
3.  Strictness: Do not give full marks unless the implementation is rigorous. "Looking similar" is not enough for a max score; it must be "functionally equivalent.\\
\\
4. Output Format
Please provide your evaluation in the following format:\\
\\
Paper Title: Title\\
Dimension: [Score](Justification (Briefly explain matches/discrepancies)\\
Method: [**/10](**)\\
Parameters: [**/5](**)\\
Pipeline: [**/5](**)\\
Total Score: **/20
}
\end{tcolorbox}
\vspace{-5pt}
\caption{Prompt for human evaluation instructions.}
\label{fig:human_eval_instruction2}
\end{figure*}

\begin{figure*}[ht]
\begin{tcolorbox}[colback=white, colframe=black, title=Prompt for summarizing 5 key points proposed in the paper]
\small\texttt{\textcolor{blue}{TASK} = MovingMnist \\
\textcolor{blue}{METRIC} = MSE \\
\textcolor{blue}{TITLE} = SimVP: Simpler yet Better Video Prediction \\
\textcolor{blue}{PATH} = bench/simvp/source \\
\textcolor{blue}{INSTRUCTION} = You are assigned an arXiv paper \{\textcolor{blue}{TITLE}\} to replicate. You need to replicate the experiment conducted for \{\textcolor{blue}{TASK}\} dataset in the paper. Training and testing datasets are under the folder \{\textcolor{blue}{PATH}\}. Code related to processing the data is provided to you to obtain the dataset. Utilise \{\textcolor{blue}{METRIC}\} as the evaluation metric. \\
\\
The provided text above is the full text of the paper. The instructions for reproducing the experiment are  \{\textcolor{blue}{INSTRUCTION}\} with the model  \{\textcolor{blue}{MODEL}\}. \\
Now you will evaluate whether the code has replicated the instructions about \{\textcolor{blue}{TASK}\} experiments in the paper. \\
Please summarise 5 key points. These 5 key points will be used to assess whether the code has completely replicated the model, methods, and experimental setting in the paper. \\
Specifically, you are preferred to use 3 key points to summarise the proposed method, 1 for hyperparameters and 1 for training setup. Do not include the dataset generation process as a point, as the dataset has been preprocessed. \\
You should regard each part of the proposed method in the paper as a separate key point. \\
If there are formulas in the paper, you need to extract them in LaTeX format and use them as the criteria for judging whether the code has been replicated. \\
Do not include some common content as the key points to be compared. Only include the key points related to the \{\textcolor{blue}{TASK}\} task. 
If all these 5 points are replicated exactly, the code will fully replicate the paper. These key points are very important and should be as detailed as possible, which could reflect the key points to reproduce the paper.}
\end{tcolorbox}
\vspace{-5pt}
\caption{Prompt for key points summarization.}
\label{fig:prompt_summarize}
\end{figure*}

\begin{figure*}[ht]
\begin{tcolorbox}[colback=white, colframe=black, title=Prompt for paper-level evaluation]
\small\texttt{Now, I'm presenting you with a generated code. You need to check whether the details of the code correspond to the key points.\\
The experiment instructions for the generated code are\\
\{\textcolor{blue}{INSTRUCTION}\}\\
You just need to consider the model, task, and dataset used in the instructions.\\
There are a total of 5 comparison points, and each point is scored from 0 to 20. A score of 20 indicates perfect reproduction, while 0 means no reproduction at all.\\
Please rate each of the 5 comparison points separately and provide the reasons.\\
Points to compare \\
\{\textcolor{blue}{Points}\} \\
Generated code \\
\{\textcolor{blue}{Generated Code}\} \\
For each scoring criterion, you need to give a score between 0-20 points. The scoring criteria are as follows: \\
Total difference: 0-2 points. \\
Unsimilar: 3-8 points. \\
Similar: 9-14 points.\\
Roughly the same: 15-18 points.\\
Same: 19-20 points.\\
You need to evaluate with a critical perspective. Don't give high scores to the mismatched content. \\
Disregard all comments, as they do not pertain to the implementation of the code.\\
The code needs to strictly correspond to the points. Any mismatched content should result in a deduction of points. \\
The scores should be presented in the form of [x/20 points]. The final score should be the sum of the scores for each point.}
\end{tcolorbox}
\vspace{-5pt}
\caption{Prompt for paper-level score. The \textcolor{blue}{Points} are the 5 key points generated by the LLM judge, and the \textcolor{blue}{Generated Code} is the code generated by LLM Agents.}
\label{fig:code_high_level}
\end{figure*}

\begin{figure*}[ht]
\begin{tcolorbox}[colback=white, colframe=black, title=Prompt for code-level evaluation]
\small\texttt{You are an expert proficient in code analysis, specialising in comparing and evaluating code structure and functionality. Now you need to judge whether the replicated code is the same as the official code. \\
Please analyse the following two code segments: the first is model-generated code, and the second is standard training code. You don't need to pay attention to the contents related to saving, printing and logging. Focus on the model itself and its training process instead. \\
Conclusion: Summarise whether the two code segments are completely equivalent in terms of model definition and Experimental Integrity, and briefly explain the significance of the scoring results. \\
First Code Segment (Standard Experiment Code): \\
\{\textcolor{blue}{Reference Code}\} \\
Second Code Segment (Model-Generated Code): \\
\{\textcolor{blue}{Generated Code}\} \\
Your task is to conduct a detailed comparison of these two code segments, focusing on the following aspects: \\
Overall Structure: \\
The overall structure of the model code includes the data flow in the forward function and the overall model structure.  \\
Only consider the model-related code, such as the encoder-decoder structure, and ignore the data loading and other irrelevant code. \\
Model Structure: \\
Are the model architectures defined in both code segments (e.g., number of layers, activation functions, input/output dimensions) completely identical? \\
Specifically compare the implementation details, such as if a spatio-temporal processing module is present in the code, including but not limited to input processing, feature extraction methods, temporal dimension handling, spatial dimension handling, connection methods (e.g., residual connections, attention mechanisms), and parameter settings. Apply this level of comparison to all modules. 
Check for any subtle differences (e.g., convolution kernel size, pooling methods, normalisation techniques). \\
Training Details: \\
Compare whether the model hyperparameters (e.g., learning rate, batch size, optimiser type, learning rate decay strategy) are consistent. Verify whether the loss function’s definition and implementation are identical, including the loss calculation formula and weight assignments. \\
Experimental Integrity: \\
Compare the implementation of the training process, including the training pipeline, data preprocessing, and gradient update logic, to determine if they are equivalent. The most crucial thing you need to pay attention to is the integrity of the experiment. Check for any functional differences (e.g., initialisation methods, early stopping mechanisms). There is no requirement for checkpoint saving, logging information and multi-GPU training. Do not consider these contents.\\
\\
Pay special attention to analysing the implementation of the module in the model structure, ensuring a detailed comparison of each layer’s specific parameters and computational logic. \\
If differences are found, clearly indicate the specific code lines or modules where they occur and analyse their potential impact on model performance or behaviour. \\
Ignore differences in code style (e.g., variable naming, comments) and focus on functionality and implementation logic.\\
You need to focus on the code implementation and don't need to consider comments and function names, and other contents irrelevant to the implementation.}
\end{tcolorbox}
\vspace{-5pt}
\caption{Prompt for code-level score. The \textcolor{blue}{Reference Code} and \textcolor{blue}{Generated Code} are our curated official implementations and Agents' generated code, respectively.}
\label{fig:code_low_level1}
\end{figure*}

\begin{figure*}[ht]
\begin{tcolorbox}[colback=white, colframe=black, title=Prompt for code-level evaluation]
\small\texttt{Scoring Criteria (Total: 100 points): \\
Overall Structure (25 points): Consistency in the overall structure of the model-related code, including the overall data flow pipeline in the forward function, and the overall modules are structured. \\
Model Details (25 points): Consistency in the implementation of the model structure. All the layers should be compared. If the names are different but the internal functions are the same, you should consider them as the same.  \\
Training Details (25 points): Consistency in hyperparameters, loss function, learning rate decay, etc. \\
Experimental Integrity (25 points): Consistency in training loop, data processing, and other pipeline. You just need to compare the training, testing pipeline included in the code. \\
\\
You need to analyse the code details and implementation before giving the score. Firstly, analyse the overall structure of the model and then analyse it specifically for each module. \\
For custom blocks, the details of how custom blocks are implemented should be analysed. \\
For some official implementation, you need to analyse whether the implementation of custom is the same as the official one based on your understanding. Ignoring the differences of programming languages and only considering the code implementation. \\
\\
For each scoring criterion, you need to give a score between 0-25 points. The scoring criteria are as follows: \\
Total difference: 0-4 points.\\
Unsimilar: 5-10 points.\\
Similar: 12-16 points.\\
Roughly the same: 17-22 points. \\
Same: 23-25 points.\\
The MLP and a single linear layer should be considered roughly the same. \\
Similarly, when the functions of two codes are the same, but the implementations are different.
Unsimilar should be when the functions of two codes are different. \\
You need to analyse the specific implementation of the code, do not just focus on the name. Make detailed evaluations based on the details. Ignore all the comments and focus only on the code. \\
The scores should be presented in the form of [x/25 points]. The final score should be the sum of the scores for each point.}
\end{tcolorbox}
\vspace{-5pt}
\caption{Prompt for code-level score. The \textcolor{blue}{Reference Code} and \textcolor{blue}{Generated Code} are our curated official implementations and Agents' generated code, respectively.}
\label{fig:code_low_level2}
\end{figure*}

\begin{figure*}[t]
\begin{tcolorbox}[colback=white, colframe=black, title=Prompt for mixed-level evaluation]
\small\texttt{You are an expert code reviewer specialising in evaluating the fidelity of research paper implementations. Your task is to meticulously compare the generated code against a set of key points from a research paper. Crucially, you will use the provided reference code as the ground truth to understand how each key point is specifically and correctly implemented. \\
The experiment instruction is
\{\textcolor{blue}{INSTRUCTION}\}
You just need to consider the model, task, and dataset used in the instructions.\\
Inputs You Will Be Provided With: \\
1. Points: A list of key concepts, mechanisms, algorithms, or architectural features from the research paper that the generated code is supposed to implement.\\
2. Reference code: The official source code accompanying the research paper. This code serves as the benchmark for understanding the precise, intended implementation details of each key point.\\
3. Generated code: The generated code that needs to be evaluated for its accuracy in reproducing the key points as they are implemented in the reference code.\\
Your Evaluation Process:\\
1. Understand Key Point via Reference Code: For each key point, first, thoroughly examine the reference code. Identify and describe the specific segments of the reference code (e.g., functions, classes, logic blocks) that implement this key point. Summarize how the reference code realizes this key point. This understanding will be your basis for comparison. \\
2. Analyse Generated Code against Reference Implementation:Now, review the generated code (generated code) to find its implementation of the same key point. Compare this implementation directly against your understanding of how it was done in the reference code. Focus on whether the logic, structure, and functional outcome are equivalent.\\
3.  Score the Replication: Based on your comparative analysis, assign a score from 0 to 20 to the generated code for its replication of this specific key point, using the scoring rubric below.\\
4. Provide Detailed Justification: Clearly articulate the reasons for your score. Specifically highlight matches and discrepancies between the generated code's implementation and the reference code's implementation of the key point. Explain why it matches or why it deviates.\\
Scoring Rubric:\\
0-2 points (Total difference): The core innovation point (as demonstrated in the reference code) is not replicated at all in the generated code, or the implementation is fundamentally flawed, missing major functional aspects, or entirely incorrect when compared to the reference code. \\
3-8 points (Unsimilar): The key point is replicated in the generated code, but not completely accurately or comprehensively when compared to the reference code. Some aspects of the reference code's implementation might be present, but there are noticeable inaccuracies, missing details, or differences in logic that might affect functionality or deviate from the paper's intended mechanism, as shown in the official code.\\
9-14 points (Similar): The key point is replicated completely and accurately in the generated code. The implementation in the generated code closely mirrors the logic, structure, and functional behavior of the corresponding implementation in the reference code, although there might be some non-critical differences or alternative approaches that achieve the same core outcome.\\
15-18 points (Roughly the same): The key point is replicated completely, accurately, and comprehensively in the generated code. The implementation is highly consistent with the reference code in logic, structure, and function, differing only in very minor, non-functional ways that do not impact the core mechanism.\\
19-20 points (Same): The implementation of the key point in the generated code is identical or functionally equivalent to the reference code, representing a code-level copy or a near-perfect replication of the relevant sections.}
\end{tcolorbox}
\vspace{-5pt}
\caption{Prompt for mixed-level score. The \textcolor{blue}{Reference Code} and \textcolor{blue}{Generated Code} are our curated official implementations and Agents' generated code, respectively.}
\label{fig:code_multi_level1}
\end{figure*}

\begin{figure*}[ht]
\begin{tcolorbox}[colback=white, colframe=black, title=Prompt for mixed-level evaluation]
\small\texttt{Critical Evaluation Guidelines:\\
Conduct your evaluation with a stringent and critical perspective. Do not award high scores for superficial similarities or implementations that do not match the essence of the reference code's approach for a given key point.\\
Your evaluation must be based solely on the executable code logic. Disregard all comments in both the reference code and the generated code as they do not pertain to the functional implementation.\\
At the same time, you need to pay attention not only to the key points themselves, but also to all the details related to them. If the key points correspond but the related implementations are different, it will still affect the reproduction effect.
Please evaluate each key point in the points with the generated code. You should both determine the overall similarity and concrete scores. For example, when you decide the two codes are similar, you should also determine the level of similarity.\\
When evaluating specific implementations, prioritise the equivalence of core structure and function over superficial differences, such as the number of modules. If generated code achieves the same functional outcome and structural design as the reference, it should be considered equivalent, irrespective of modular composition.\\
\\
Begin Evaluation:\\
Points to compare:\{\textcolor{blue}{points}\}\\
Reference Code:\{\textcolor{blue}{Reference Code}\}\\
Generated code:\{\textcolor{blue}{Generated Code}\}\\
\\
Output Format:\\
For each key point, please provide:
1.  Key Point:*[Name/Description of the key point being evaluated]\\
2.  Reference Code Implementation Summary:*[Your summary of how this key point is implemented in the reference code]\\
3.  Generated Code Analysis \& Comparison:*[Your detailed analysis of the generated code's attempt to implement this point, comparing it directly to the reference code's approach]\\
4.  Score:*[x/20 points]\\
5.  Reasoning for Score:*[Detailed justification based on the comparison]\\
Sum the overall scores for each key point to provide a final score out of 100 points, and include a summary of the overall evaluation.\\
Overall Score:*[x/100 points]\\}
\end{tcolorbox}
\vspace{-5pt}
\caption{Prompt for mixed-level score. The \textcolor{blue}{Reference Code} and \textcolor{blue}{Generated Code} are our curated official implementations and Agents' generated code, respectively.}
\label{fig:code_multi_level2}
\end{figure*}

\end{document}